\documentclass{svmult}
\pdfoutput=1

\usepackage{makeidx}         
\usepackage{graphicx}        
\usepackage{multicol}        
\usepackage[bottom]{footmisc}
\usepackage{amsmath}

\makeindex             

\makeatletter 
\long\def\@makecaption#1#2{%
  \vskip\abovecaptionskip
  \sbox\@tempboxa{#1: #2}%
  \ifdim \wd\@tempboxa >\hsize
    #1: #2\par
  \else
    \global \@minipagefalse
    \hb@xt@\hsize{\box\@tempboxa\hfil}%
  \fi
  \vskip\belowcaptionskip}
\makeatother

\begin{document}

\title*{Using Genetic Algorithms to Optimise Rough Set Partition Sizes for HIV Data Analysis}
\titlerunning{GA's to Optimise Rough Sets}
\author{Bodie Crossingham and Tshilidzi Marwala}
\institute{School of Electrical and Information Engineering,
University of the Witwatersrand \texttt{t.marwala@ee.wits.ac.za}}
%
%
\maketitle

\begin{abstract}

In this paper, we present a method to optimise rough set partition
sizes, to which rule extraction is performed on HIV data. The
genetic algorithm optimisation technique is used to determine the
partition sizes of a rough set in order to maximise the rough sets
prediction accuracy. The proposed method is tested on a set of
demographic properties of individuals obtained from the South
African antenatal survey. Six demographic variables were used in
the analysis, these variables are; race, age of mother, education,
gravidity, parity, and age of father, with the outcome or decision
being either HIV positive or negative. Rough set theory is chosen
based on the fact that it is easy to interpret the extracted
rules. The prediction accuracy of equal width bin partitioning is
57.7\% while the accuracy achieved after optimising the partitions
is 72.8\%. Several other methods have been used to analyse the HIV
data and their results are stated and compared to that of rough
set theory (RST).

\end{abstract}

\section{Introduction}
\label{sec:1}

In the last 20 years, over 60 million people have been infected
with HIV (Human immunodeficiency virus), and of those cases, 95\%
are in developing countries~\cite{Lasry}. In 2006 alone, an
estimated 39.5 million people around the world were living with
HIV, with 27.5 million of those people living in Sub-Saharan
Africa. During this year, AIDS (Acquired Immune Deficiency
Syndrome) claimed an estimated 2.9 million lives~\cite{COWUN}. HIV
has been identified as the cause of AIDS. The effect of AIDS is
not only detrimental to the individual infected but has a
devastating effect on the economic, social, security and
demographic levels of a country. Because AIDS is killing people in
the prime of their working and parenting lives, it represents a
grave threat to economic development. In the worst affected
countries, the epidemic has already reversed many of the
development achievements of the past generation~\cite{COWUN}.
There are many other negative economic effects of AIDS, it has a
large negative impact on the social and security levels of a
country. Social levels drop as the health and educational
development, that is supposed to benefit poor people, is impeded
as well as the average life expectancy drops. It is estimated that
by 2010 the number of orphans is expected to double from that in
2006~\cite{COWUN}.

Early studies on HIV/AIDS focused on the individual
characteristics and behaviours in determining HIV risk, Fee and
Krieger refer to this as \emph{biomedical
individualism}~\cite{Fee}. But it has been determined that the
study of the distribution of health outcomes and their social
determinants is of more importance, this is referred to as
\emph{social epidemiology}~\cite{Poundstone}. This study uses
individual characteristics as well as social and demographic
factors in determining the risk of HIV.

It is thus evident from above that the analysis of HIV is of the
utmost importance. By correctly forecasting HIV, the causal
interpretations of a patients being seropositive (infected by HIV)
is made much easier. Previously, computational intelligence
techniques have been used extensively to analyse HIV. Leke
\emph{et al} have used autoencoder network classifiers, inverse
neural networks, as well as conventional feedforward neural
networks to analyse HIV~\cite{Leke:autoenc,Tim,Leke:inverse}, they
used the inverse neural network for adaptive control of HIV status
to understand how the demographic factors affect the risk of HIV
infections~\cite{Leke:inverse}.

Although an accuracy of 92\% is achieved when using the
autoencoder method~\cite{Leke:autoenc}, it is disadvantageous due
to its ``black box'' nature, this also applies to the other
mentioned neural network techniques. Neural network connection
weights and transfer functions are frozen upon completion of
training of the neural network~\cite{Cannady}. Neural networks
offer accuracy over analysis of data, but in the case of analysing
HIV data, it can be argued that interpretability of the data is of
more importance than just prediction. It is due to this fact that
rough set theory (RST) is proposed to forecast and interpret the
causal effects of HIV.

Rough sets have been used in various biomedical
applications~\cite{Ohrn,Rowland,Peszek}, other applications of RST
include the prediction of aircraft component failure, fault
diagnosis and stock market analysis~\cite{Famili,Tay,Golan}. But
in most applications, RST is used primarily for prediction.
Rowland \emph{et al} compared the use of RST and neural networks
for the prediction of ambulation spinal cord
injury~\cite{Rowland:spinal}, and although the neural network
method produced more accurate results, its ``black box'' nature
makes it impractical for the use of rule extraction problems.

Poundstone \emph{et al} related demographic properties to the
spread of HIV. In their work they justified the use of demographic
properties to create a model to predict HIV from a given database,
as is done in this study. RST uses the social and demographic
factors to predict HIV status, this in turn provides insight into
which variables are most sensitive in determining HIV status. For
example, if 90\% of HIV positive cases have limited and/or no
education, whereas 85\% of HIV negative cases have at least
secondary school education, this would clearly indicate that by
improving the nations education, the percentage of seropositive
patients should decrease.

In order to achieve the best accuracy, the rough set partitions or
discretisation process needs to be optimised. The optimisation is
done by a genetic algorithm (GA), where the fitness function aims
to achieve the highest accuracy produced by the rough set.
Literature reviews have shown that limited work has been done on
the optimisation of rough set partition sizes.

The background of the topic is stated in section~\ref{sec:2}, a
discussion on rough set theory and the formulation of the rough
sets from which rules are extracted are given in
section~\ref{sec:3}. Section~\ref{sec:4} explains how the genetic
algorithm is used to optimise the rough set partitions, and then
in section~\ref{sec:5} the results obtained for partitioning the
data using equal width bin are compared to that of the results
obtained when optimising the partition sizes using a GA.

\section{Background}
\label{sec:2}

Rough set theory was introduced by Zdzislaw Pawlak in the early
1980s~\cite{Pawlak:Book}. RST is a mathematical tool which deals
with vagueness and uncertainty. It is of fundamental importance to
artificial intelligence (AI) and cognitive science and is highly
applicable to this study performing the task of machine learning
and decision analysis. Rough sets are useful in the analysis of
decisions in which there are inconsistencies. To cope with these
inconsistencies, lower and upper approximations of decision
classes are defined~\cite{Inuiguchi}. Rough set theory is often
contrasted to compete with fuzzy set theory (FST), but it in fact
complements it~\cite{Pawlak:Book}. One of the advantages of RST is
it does not require \emph{a priori} knowledge about the data set,
and it is for this reason that statistical methods are not
sufficient for determining the relationship between the
demographic variables and their respective outcomes.

The data set used in this paper was obtained from the South
African antenatal sero-prevalence survey of 2001. The data was
obtained through questionnaires completed by pregnant women
attending selected public clinics and was conducted concurrently
across all nine provinces in South Africa. The sentinel population
for the study only included pregnant women attending an antenatal
clinic for the first time during their current pregnancy. The
choice of the first antenatal visit is made to minimise the chance
for one woman attending two clinics and being included in the
study more than once~\cite{HIV:Survey}.

The six demographic variables considered are: \emph{race, age of
mother, education, gravidity, parity} and, \emph{age of father},
with the outcome or decision being either HIV positive or
negative.

The HIV status is the decision represented in binary form as
either a 0 or 1, with a 0 representing HIV negative and a 1
representing HIV positive. The input data was discretised into
four partitions. This number was chosen as is gave a good balance
between computational efficiency and accuracy. The race attribute
is presented on a scale 1 to 4, where the numbers represent White,
African, Coloured and Asian respectively. The parents ages are
given and discretised accordingly, education is given as an
integer, where 13 is the highest level of education, indicating
tertiary education. Gravidity is defined as the number of times
that a woman has been pregnant, whereas parity is defined as the
number of times that she has given birth. It must be noted that
multiple births during a pregnancy are indicated with a parity of
one. Gravidity and parity also provide a good indication of the
reproductive health of pregnant women in South Africa.

\section{Rough Set Theory and Rough Set Formulation}
\label{sec:3}

Rough set theory deals with the approximation of sets that are
dif\mbox{}ficult to describe with the available
information~\cite{Rowland}. It deals predominantly with the
classification of imprecise, uncertain or incomplete information.
Some concepts that are fundamental to RST theory are given below.

\subsection{Information Table}
\label{subsec:1}

The data is represented using an information table, an example for
the HIV data set for the \emph{ith} object is given below:

\begin{table}[ht]
\caption{Information Table of the HIV Data.}
\centering
\begin{tabular}{|c|c|c|c|c|c|c|c|c|}
  \hline
   & Race & Mothers Age & Education & Gravidity & Parity & Fathers Age && HIV Status \\
  \hline
  $Obj^{(1)}$ & 2 & 32 & 13 & 1 & 1 & 22 && 1 \\
  $Obj^{(2)}$ & 3 & 22 & 5 & 2 & 1 & 25 && 1 \\
  $Obj^{(3)}$ & 1 & 35 & 6 & 1 & 0 & 33 && 0 \\
  . & . & . & . & . & . & . && . \\
  $Obj^{(i)}$ & 2 & 27 & 9 & 3 & 2 & 30 && 0 \\
  \hline
\end{tabular}
\end{table}

In the information table, each row represents a new case (or
\emph{object}). Besides \emph{HIV Status}, each of the columns
represent the respective case's variables (or \emph{condition
attributes}). The \emph{HIV Status} is the outcome (also called
the \emph{concept} or \emph{decision attribute}) of each object.
The outcome contains either a 1 or 0, and this indicates whether
the particular case is infected with HIV or not.

\subsection{Information System}
\label{subsec:2}

Once the information table is obtained, the data is discretised
into four partitions as mentioned earlier. An information system
can be understood by a pair
\begin{math}\Lambda\end{math} = (\textbf{U},A), where \textbf{U} and A, are finite, non-empty
sets called the universe, and the set of attributes,
respectively~\cite{Peszek}.

For every attribute \emph{a $\in$ A}, we associate a set
\emph{$V_{a}$}, of its values, where \emph{$V_{a}$} is called the
value set of \emph{a}.

\begin{equation} \label{valueSet}
    \emph{a}:\textbf{U}\rightarrow\emph{$V_{a}$}
\end{equation}

Any subset \emph{B} of \emph{A} determines a binary relation
\emph{I(B)} on \textbf{U}, which is called an indiscernibility
relation. This concept will be explained below.

\subsection{Indiscernibility Relation}
\label{subsec:3}

The main concept of rough set theory is an indiscernibility
relation (indiscernibility meaning indistinguishable from one
another). Sets that are indiscernible are called elementary sets,
and these are considered the building blocks of RST's knowledge of
reality. A union of elementary sets is called a crisp set, while
any other sets are referred to as rough or vague.

More formally, for a given information system
\begin{math}\Lambda\end{math}, then for any subset
\emph{B}$\subseteq$\emph{A}, there is an associated equivalence
relation \emph{I(B)} called the \emph{B-indiscernibility relation}
and is represented as shown in \ref{indiscern} below:

\begin{equation} \label{indiscern}
    (\emph{x},\emph{y}) \emph{ $\in$ } \emph{I(B) iff }  a(x)=a(y)
\end{equation}

RST offers a tool to deal with indiscernibility, the way in which
it works is, for each concept/decision X, the greatest definable
set containing X and the least definable set containing X are
computed. These two sets are called the lower and upper
approximation respectively.

\subsection{Lower and Upper Approximations}
\label{subsec:4}

The sets of cases/objects with the same outcome variable are
assembled together. This is done by looking at the ``purity" of
the particular objects attributes in relation to its outcome. In
most cases it is not possible to define cases into crisp sets, in
such instances lower and upper approximation sets are defined.

The lower approximation is defined as \emph{the collection of
cases whose equivalence classes are fully contained in the set of
cases we want to approximate}~\cite{Rowland}. The lower
approximation of set \emph{X} is denoted \emph{$\underline{B}$X}
and mathematically it is represented as:

\begin{equation} \label{lower}
    \emph{$\underline{B}$X} = \{{\emph{x $\in$ }} \textbf{U}:
    \emph{B}(\emph{x})\emph{$\subseteq$X}\}
\end{equation}

The upper approximation is defined as \emph{the collection of
cases whose equivalence classes are at least partially contained
in the set of cases we want to approximate}~\cite{Rowland}. The
upper approximation of set \emph{X} is denoted
\emph{$\overline{B}$X} and is mathematically represented as:

\begin{equation} \label{upper}
    \emph{$\overline{B}$X} = \{{\emph{x $\in$ }} \textbf{U}:
    \emph{B}(\emph{x})\emph{$\cap$X$ \not= \emptyset$}\}
\end{equation}

It is through these lower and upper approximations that any rough
set is defined. Lower and upper approximations are defined
differently in literature, but it follows that a crisp set is only
defined for \emph{$\overline{B}$X} = \emph{$\underline{B}$X}.

It must be noted that for most cases in RST, reducts are generated
to enable us to discard functionally redundant
information~\cite{Pawlak:Book}. And although reducts are one of
the main advantages of RST, it is ignored for the purpose of this
paper, i.e. the optimisation of discretised partitions.

\subsection{Rough Membership Function}
\label{subsec:5}

The rough membership function is described; $\mathrm{\mu}_A^X :
U\rightarrow [0,1]$ that, when applied to object x, quantifies the
degree of relative overlap between the set X and the
indiscernibility set to which x belongs. This membership function
is a measure of the plausibility of which an object x belongs to
set X. This membership function is defined as:

\begin{equation} \label{memFunc}
    \mathrm{\mu}_A^X = \frac{|[X]_B \cap X|}{[X]_B}
\end{equation}

\subsection{Rough Set Accuracy}
\label{subsec:6}

The accuracy of rough sets provides a measure of how closely the
rough set is approximating the target set. It is defined as the
ratio of the number of objects which can be positively placed in X
to the number of objects that can be possibly be placed in X. In
other words it is defined as the number of cases in the lower
approximation, divided by the number of cases in the upper
approximation; $0 \leq \alpha_p (X)\leq 1$

\begin{equation} \label{accu}
    \alpha_p (X) = \frac{|\emph{$\underline{B}$X}|}{|\emph{$\overline{B}$X}|}
\end{equation}

\subsection{Rough Sets Formulation}
\label{subsec:7}

The process of modelling the rough set can be broken down into five stages;\\

 The first stage would be to select the data. The data
to be used is obtained from the South African antenatal survey of
2001~\cite{HIV:Survey}.

The second stage involves pre-processing the data to ensure it is
ready for analysis, this stage involves discretising the data and
removing unnecessary data (cleaning the data). Although the
optimal selection of set sizes for the discretisation of
attributes will not be known at first, an optimisation technique
(genetic algorithm) will be run on the set to ensure the highest
degree of accuracy when forecasting outcomes. This will be
explained more clearly below and is illustrated in
figure~\ref{Block}.

If reducts were considered, the third stage would be to use the
cleaned data to generate reducts. A reduct is the most concise way
in which we can discern object classes~\cite{Witlox}. In other
words, \emph{a reduct is the minimal subset of attributes that
enables the same classification of elements of the universe as the
whole set of attributes}~\cite{Pawlak:Book}. To cope with
inconsistencies, lower and upper approximations of decision
classes are
defined~\cite{Ohrn,Peszek,Pawlak:Book,Inuiguchi,Witlox}.

Stage four is where the rules are extracted or generated. The
rules are normally determined based on condition attributes
values~\cite{Law}. Once the rules are extracted, they can be
presented in an \emph{if} CONDITION(S)-\emph{then} DECISION
format~\cite{Brain:Phd}.

The final or fifth stage involves testing the newly created rules
on a test set. The accuracy will be noted and sent back into the
genetic algorithm in step two and the process will continue until
the optimum or highest accuracy is achieved.

\paragraph{Pre-processing Data}
As with many surveys, there is missing and/or incorrect data. This
data needs to be cleaned before any processing can be performed on
it. The first irregularity would be the case of missing data. This
could be due to the fact that surveyees may have omitted certain
information, it could also be attributed to the errors being made
when the data was entered onto the computer. Such cases are
removed from the data set. The second irregularity would be
information that is false. Such an instance would be if gravidity
was zero and parity was at least one. Gravidity is defined as the
number of times that a woman has been pregnant, and parity is
defined as the number of times that she has given birth. Therefore
it is impossible for a woman to have given birth, given she has
not been pregnant, such cases are removed from the data set. As
mentioned earlier, multiple births are still indicated with a
parity of one, therefore if parity is greater than gravidity, that
particular case is also removed from the data set. Only 12945
cases remained from a total of 13087.

\paragraph{Rule Extraction}
Once RST was applied to the HIV data, 329 unique distinguishable
cases and 123 indiscernible cases were extracted. From the data
set of 12945 cases, the data is only a representative of 452 cases
out of the possible 4096 unique combinations. From~\ref{accu} the
accuracy of the rough set is calculated to be 72.8\%. The 329
cases of the lower approximation are rules that always hold, or
are definite cases. The 123 cases of the upper approximation can
only be stated with a certain plausibility. Examples of both cases
are stated below:

\subparagraph{Lower Approximation Rules}
\begin{enumerate}
    \item \textbf{If} Race = African \textbf{and} Mothers Age = 23 \textbf{and}
        Education = 4 \textbf{and} Gravidity = 2 \textbf{and} Parity
        = 1 \textbf{and} Fathers Age = 20 \textbf{Then} HIV = Most
        Probably Positive
    \item \textbf{If} Race = Asian \textbf{and} Mothers Age = 30 \textbf{and}
        Education = 13 \textbf{and} Gravidity = 1 \textbf{and} Parity
        = 1 \textbf{and} Fathers Age = 33 \textbf{Then} HIV = Most
        Probably Negative

 \end{enumerate}

\subparagraph{Upper Approximation Rules}
\begin{enumerate}
    \item \textbf{If} Race = Coloured \textbf{and} Mothers Age = 33 \textbf{and}
        Education = 7 \textbf{and} Gravidity = 1 \textbf{and} Parity
        = 1 \textbf{and} Fathers Age = 30 \textbf{Then} HIV = Positive with
        plausibility = 0.33333
    \item \textbf{If} Race = White \textbf{and} Mothers Age = 20 \textbf{and}
        Education = 5 \textbf{and} Gravidity = 2 \textbf{and} Parity
        = 1 \textbf{and} Fathers Age = 20 \textbf{Then} HIV = Positive with
        plausibility = 0.06666
\end{enumerate}

\section{Genetic Algorithm}
\label{sec:4}

\begin{figure} [htp]
    \centering
    \includegraphics[width=1\textwidth]{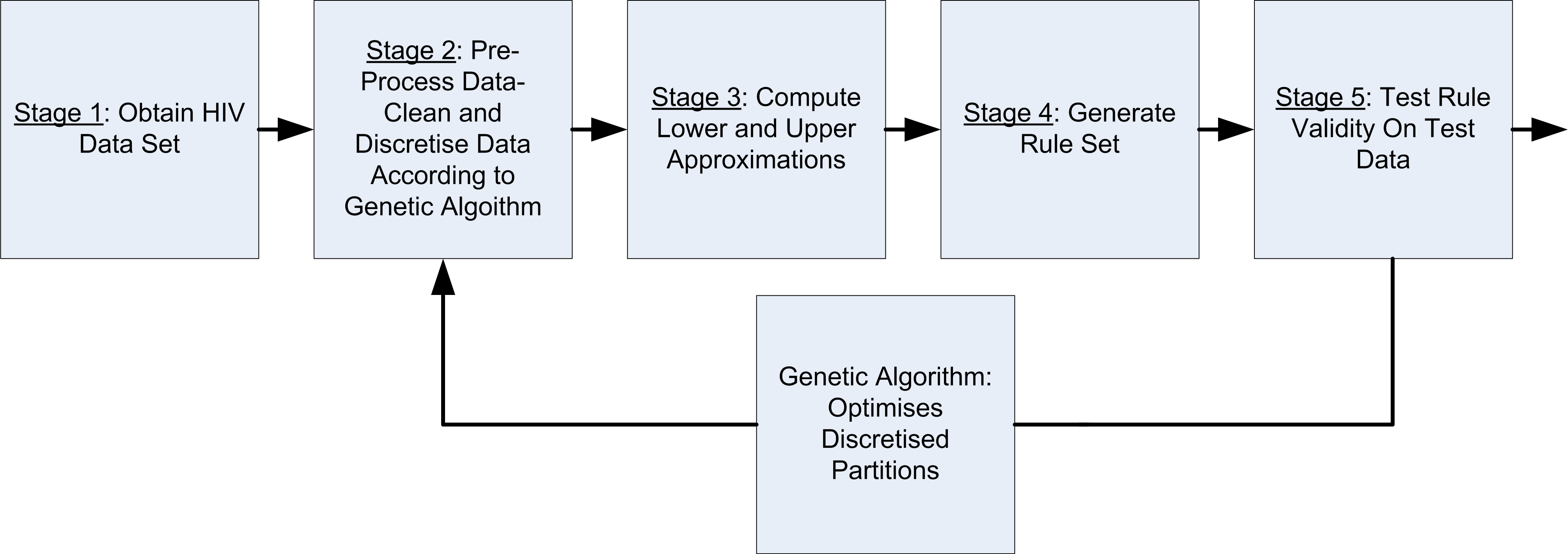}
    \caption{Block Diagram of the Sequence of Events in Modelling
    MID} \label{Block}
\end{figure}

A genetic algorithm (GA) is a stochastic search procedure for
combinatorial optimisation problems based on the mechanism of
natural selection~\cite{Malve}. Genetic algorithms are a
particular class of evolutionary algorithms that use techniques
inspired by evolutionary biology such as inheritance, mutation,
selection, and crossover. The fitness/evaluation function is the
only part of the GA that has any knowledge about the problem. The
fitness function tries to maximise the accuracy of the rough set.
Figure~\ref{Block} illustrates the process of computing the rough
sets simultaneously with the GA optimising the partition sizes.

The pseudo-code algorithm for genetic algorithms is given below;
\begin{enumerate}
    \item Initialise a population of chromosomes
    \item Evaluate each chromosome (individual) in the population
    \begin{enumerate}
        \item Create new chromosomes by mating chromosomes in the
                current population (using crossover and mutation)
        \item Delete members of the existing population to make way for
                the new members
        \item Evaluate the new members and insert them into the
                population
    \end{enumerate}
    \item Repeat stage 2 until some termination condition is
    reached, in this case until 100 generations were reached.
    \item Return the best chromosome as the solution
\end{enumerate}

As selection functions, mutation and crossover functions are
relevant to each specific problem, for this purpose of this paper,
the best results were obtained using normal geometric selection, a
uniform mutation and cyclic crossover, an initial population of 20
individuals was chosen. GAs also may prematurely converge to a
local minimum, but they do incorporate a diversification mechanism
to avoid this, the mechanism used is mutation.

\section{Results Obtained}
\label{sec:5}

The accuracy of the rough set was calculated for two cases, the
first case was for when the partitions were discretised manually
into equal width bins, and the second case was when the partition
sizes were chosen optimally by implementing a GA. The first case
yielded 225 cases, of which there were 130 unique discernible
cases and 95 indiscernible cases. This represents an accuracy of
57.7\%. The second case, yielded 452 cases. 329 of the cases were
discernible while 123 were indiscernible. This produced an
accuracy of 72.8\%. The results are clearly better for the
optimised case. As a result of implementing RST on the data set,
the rules extracted are explicit and easily interpreted. RST will
however compromise accuracy over rule interpretability, and this
is brought about in the discretisation process where the
granularity of the variables are decreased.

\section{Conclusion}
\label{sec:6}

A genetic algorithm was successfully applied to RST on the HIV
data set. Although RST does not produce accuracies as high as
those of other previous computational intelligence methods, it
does however produce explicit and easy-to-interpret rules. An
accuracy of 72.8\% was produced by the rough set when applied to
the HIV data set. The GA optimisation method produced good results
but GAs may prematurely converge towards local optima.
Recommendations for future work include the application of other
optimisation techniques such as particle swarm optimisation (PSO).
PSO is advantageous over GAs as it is easy to implement and there
are fewer parameters to adjust. Different divergence mechanisms
such as elitism can also be considered for a possible increase in
accuracy.

\bibliographystyle{witseie}
\bibliography{biblio}

\begin{thebibliography}{10}
\newcommand{\enquote}[1]{``#1''}
\providecommand{\url}[1]{{\tt #1}}
\providecommand{\urlprefix}{URL }

\bibitem{Lasry}
A.~Lasry, G.~S. Zaric, and M.~W. Carter.
\newblock \enquote{Multi-level resource allocation for HIV prevention: A model
  for developing countries.}
\newblock {\em European Journal of Operational Research\/}, vol. 180, p.
  786–799, 2007.

\bibitem{COWUN}
\enquote{UNAIDS.}
\newblock www.unaids.org/en/HIV\_data/2006GlobalReport/default.asp/. Last
  accessed: 20/3/2007.

\bibitem{Fee}
E.~Fee and N.~Krieger.
\newblock \enquote{Understanding AIDS: historical interpretations and the
  limits of biomedical individualism.}
\newblock {\em American Journal of Public Health\/}, vol.~83, pp. 1477--1486,
  1993.

\bibitem{Poundstone}
K.~E. Poundstone, S.~A. Strathdee, and D.~D. Celentano.
\newblock \enquote{The Social Epidemiology of Human Immunodeficiency
  Virus/Acquired Immunodeficiency Syndrome.}
\newblock {\em Epidemiol Reviews\/}, vol.~26, pp. 22--35, 2004.

\bibitem{Leke:autoenc}
B.~B. Leke, T.~Marwala, and T.~Tettey.
\newblock \enquote{Autoencoder networks for HIV classification.}
\newblock {\em Current Science\/}, vol.~91, pp. 1467--1473, 2006.

\bibitem{Tim}
B.~B. Leke, T.~Marwala, T.~Tim, and M.~Lagazio.
\newblock \enquote{Prediction of HIV Status from Demographic Data Using Neural
  Networks.}
\newblock In {\em Proceedings of the IEEE International Conference on Systems,
  Man and Cybernetics\/}, pp. 2339--2344. Taiwan 2006.

\bibitem{Leke:inverse}
T.~Marwala, T.~Tettey, and B.~B. Leke.
\newblock \enquote{Using inverse neural network for HIV adaptive control.}
\newblock {\em International Journal of Computational Intelligence Research\/},
  vol.~3, pp. 11--15, 2007.

\bibitem{Cannady}
J.~Cannady.
\newblock \enquote{Artificial Neural Networks for Misuse Detection.}
\newblock In {\em Proceedings of the 1998 National Information Systems Security
  Conference\/}. Arlington, VA 1998.

\bibitem{Ohrn}
A.~Ohrn.
\newblock \enquote{Discernibility and Rough Sets in Medicine: Tools and
  Applications.}, 1999.
\newblock PhD Thesis, Department of Computer and Information Science Norwegian
  University of Science and Technology.

\bibitem{Rowland}
A.~Ohrn and T.~Rowland.
\newblock \enquote{Rough Sets: A Knowledge Discovery Technique for
  Multifactorial Medical Outcomes.}
\newblock {\em American Journal of Physical Medicine and Rehabilitation\/},
  vol.~79, pp. 100--108, 2000.

\bibitem{Peszek}
A.~Deja and P.~Peszek.
\newblock \enquote{Applying Rough Set Theory to Multi Stage Medical
  Diagnosing.}
\newblock {\em Fundamenta Informaticae\/}, , no.~4, pp. 387--408, 54 2003.

\bibitem{Famili}
J.~Pe-a, S.~Létourneau, and A.~Famili.
\newblock \enquote{Application of Rough Sets Algorithms to Prediction of
  Aircraft Component Failure.}
\newblock In {\em Proceedings of the Third International Symposium on
  Intelligent Data Analysis\/}. Amsterdam 1999.

\bibitem{Tay}
F.~E.~H. Tay and L.~Shen.
\newblock \enquote{Fault diagnosis based on Rough Set Theory.}
\newblock {\em Engineering Applications of Artificial Intelligence\/}, vol.~16,
  p. 39–43, 2003.

\bibitem{Golan}
R.~H. Golan and W.~Ziarko.
\newblock \enquote{A methodology for stock market analysis utilizing rough set
  theory.}
\newblock In {\em Proceedings of Computational Intelligence for Financial
  Engineering\/}, pp. 32--40. New York, USA 1995.

\bibitem{Rowland:spinal}
T.~Rowland, Ohno-Machado, and A.~Ohrn.
\newblock \enquote{Comparison of multiple prediction models for ambulation
  following spinal cord injury.}
\newblock {\em In Chute\/}, vol.~31, pp. 528--532, 1998.

\bibitem{Pawlak:Book}
Z.~Pawlak.
\newblock {\em Rough Sets, Theoretical Aspects of Reasoning about Data\/},
  chap.~3, p.~33.
\newblock Kluwer Academic Publishers, 1991.

\bibitem{Inuiguchi}
M.~Inuiguchi and T.~Miyajima.
\newblock \enquote{Rough set based rule induction from two decision tables.}
\newblock {\em European Journal of Operational Research\/}, vol. In Press,
  Corrected Proof, 2006.

\bibitem{HIV:Survey}
R.~Department~of Health.
\newblock \enquote{National HIV and Syphilis Sero-Prevalence Survey of Women
  Attending Public Antenatal Clinics in South Africa.}
\newblock http://www.info.gov.za/otherdocs/2002/hivsurvey01.pdf, 2001.

\bibitem{Witlox}
F.~Witlox and H.~Tindemans.
\newblock \enquote{The application of rough sets analysis in activity-based
  modelling. Opportunities and constraints.}
\newblock {\em Expert Systems with Applications\/}, vol.~27, p. 585–592, 2004.

\bibitem{Law}
C.~Goh and R.~Law.
\newblock \enquote{Incorporating the rough sets theory into travel demand
  analysis.}
\newblock {\em Tourism Management\/}, vol.~24, p. 511–517, 2003.

\bibitem{Brain:Phd}
B.~B. Leke.
\newblock {\em Computational Intelligence for Modelling HIV\/}.
\newblock Ph.D. thesis, University of the Witwatersrand, School of Electrical
  and Information Engineering, 2007.

\bibitem{Malve}
S.~Malve and R.~Uzsoy.
\newblock \enquote{A genetic algorithm for minimizing maximum lateness on
  parallel identical batch processing machines with dynamic job arrivals and
  incompatible job families.}
\newblock {\em Computers and Operations Research\/}, vol.~34, p. 3016–3028,
  2007.

\end{thebibliography}

\end{document}